\documentclass{article}

\usepackage{PRIMEarxiv}
\usepackage[numbers]{natbib}

\usepackage[utf8]{inputenc} 
\usepackage[T1]{fontenc}    
\usepackage{hyperref}       
\usepackage{url}            
\usepackage{booktabs}       
\usepackage{amsfonts}       
\usepackage{nicefrac}       
\usepackage{microtype}      
\usepackage{lipsum}
\usepackage{fancyhdr}       
\usepackage{graphicx}       
\graphicspath{{media/}}     

\title{Probing TryOnGAN}

\author{Saurabh Kumar \\
       DTU, Delhi, India\\
       \And 
       Nishant Sinha \\
       OffNote Labs, Bengaluru, India
}

\begin{document}

\maketitle
\begin{abstract}
    TryOnGAN is a recent virtual try-on approach, which generates highly realistic images and outperforms most previous approaches.
    In this article, we reproduce the TryOnGAN implementation and probe it along diverse angles: impact of transfer learning, variants of conditioning image generation with poses and properties of latent space interpolation. Some of these facets have never been explored in literature earlier.
    We find that transfer helps training initially but gains are lost as models train longer and pose conditioning via concatenation performs better. The latent space self-disentangles the pose and the style features and enables style transfer across poses. Our code and models are available in open source~\cite{toganrepo}.
    
\end{abstract}


\section{Introduction}
Due to recent constraints on physical retail stores, several retailers are rushing to enable consumers to try-on clothing virtually.
The goal of image-based virtual try-on (ViTON) problem is to transfer a specific clothing item (say, worn by a person $S$) to a target person $T$, where pose of $T$ and $S$ may differ.
The ViTON problem has been studied widely and the literature is vast~\cite{cpviton,adgan,togan}.
Many solutions adopt the {\it warp-and-fuse} approach: first {\it warp} the garment from $S$ to align with the target person $T$'s pose and then {\it fuse} the warped garment with the image of $T$.

A recent approach, TryOnGAN~\cite{togan}, generates highly realistic images which outperform many previous approaches.
In contrast to the widely used {\it warp-and-fuse} approaches, the TryOnGAN framework adopts a different approach: they employ the StyleGAN2 model~\cite{stylegan2}, conditioned with user-given pose embeddings as the main input and latent style embeddings as lateral inputs, to generate the desired photo-realistic images. 
To generate the try-on image, the pose embedding from person $P$ together with inverted~\cite{togan} style embedding from person $S$ is fed into a pretrained StyleGAN2 model.
Although capable of generating photo-realistic images, the StyleGAN2 model is resource-hungry and training may take several days to converge.

In this paper, we probe the design of the TryOnGAN framework along multiple dimensions: (a) how does conditional pose encoding impact generation? (b) does transfer learning from pre-trained StyleGANs enable faster convergence? (c) what are the properties of the learned latent spaces?

\section{Background: TryOnGAN}
We describe TryOnGAN~\cite{togan} briefly here.
The original StyleGAN2~\cite{stylegan2} generates RGB images conditioned by a learned constant (head) input and a set of (lateral) {\it style} vector inputs.
The generator consists of a sequence of style blocks, each taking in a latent vector $w$ that controls the style of the image generated.
The style vectors are further parameterized by a style generator network.
The first style block takes in a learned constant matrix of dimension 4x4x512. 
The TryOnGAN model~\cite{togan} prepends a {\it pose encoder} to the StyleGAN2 generator, which takes in the person pose keypoints and outputs of size 4x4, 8x8, 16x16 and 64x64. The 4x4 matrix serves as the head input to StyleGAN2, while the other matrices are concatenated to the outputs of successive style blocks. 
The StyleGAN2 discriminator is modified similarly.
This amounts to pose-conditioning the model such that the output person image always has the input pose. 
They also add a segmentation map branch to each style block. 
This modified model is trained with human images using GAN losses~\cite{gan,stylegan2}. To achieve garment transfer, TryOnGAN inverts both the target person image $P$ and the source person image $S$ to their respective latent vectors and learns to interpolate the two latents.
They claim that a linear combination of the two latent vectors can give a latent vector such that the image generated using it would have the desired garment of $(I_g)$ and the identity and other garments of $(I_p)$.

\section{Probing TryOnGAN}
The TryOnGAN~\cite{togan} model implementation is not open-source; we reproduce the implementation based on their paper.
We use Fréchet Inception Distance (FID) scores~\cite{frechet} as the main quantitative comparison metric in the paper; FID scores represent photorealism and lower FID scores are better.
We also compare images qualitatively (use image grids) for different model/training strategies.
Our code and models are available in open source~\cite{toganrepo}.

\noindent {\bf Dataset.} The TryOnGAN paper uses a private dataset with 104k 512x512 images~\cite{togan}.
To reproduce results, we use 48k 256x256 people images from the DeepFashion In-shop Clothes Retrieval Benchmark~\cite{deepfashion} . 
Following \cite{stylegan2} we report the total number of real images seen by the GAN to indicate training time. We train until model has seen 2 million (2m) images -- we observe that the best models reach FID score 10 at that point. 
Further training may further reduce FID scores; we observed that training on 2m images is sufficient to obtain images with discernable garment shapes and colors.
All experiments are done on a single NVIDIA Tesla T4 16GB GPU.

\noindent {\bf Transfer Learning.} For evaluating the impact of transfer learning, we use a StyleGAN2 model pretrained on FFHQ dataset~\cite{ffhq} and compare with training initialized with random weights (x-scratch).
We consider both pose-conditioned (PC-x) and unconditioned models (UC-x, without pose encoder).
For pose-conditioned models (PC-x), we experiment with two ways of adding pose information as input. 
In PC-concat-x models, we {\it concatenate} pose embeddings with output of style blocks whereas in PC-add-x models, we {\it add} them. 
In PC-concat-x, this changes the channel sizes in the style block, so we don’t transfer pretrained weights.

\begin{figure}[h]
\centering
\includegraphics[scale=0.15]{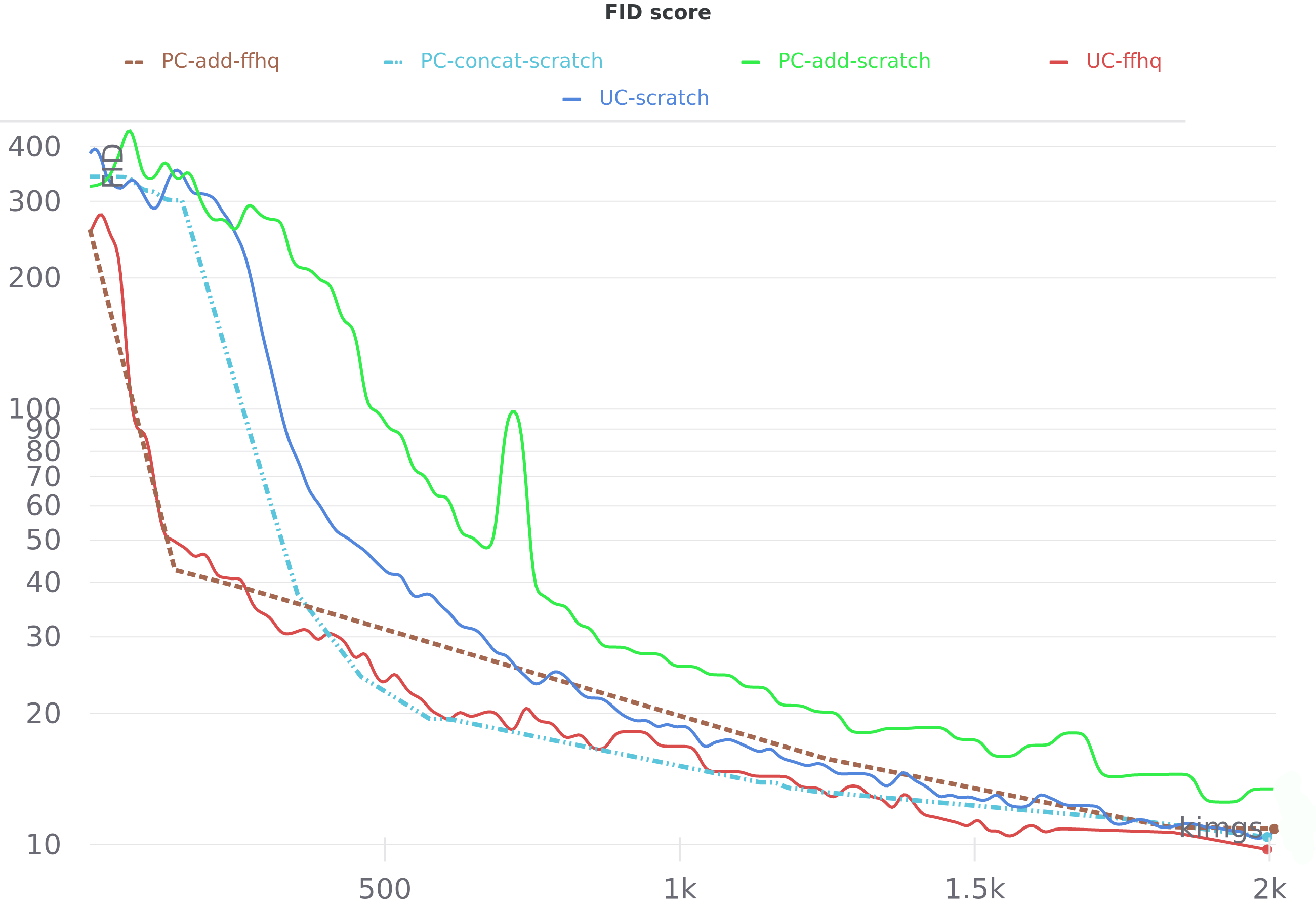}
\caption{FID scores}
\label{fig:transfer}
\end{figure}

Fig.~\ref{fig:transfer} shows the effect of transfer learning on pose-conditioned (PC-x) and unconditioned models (UC-x). 
We observe that, both in PC-x and UC-x, pretrained weights lead to a quick drop in FID scores initially, and smaller training times. 
For UC-x models, the initial FID scores drop faster for UC-ffhq vs UC-scratch.
Although the gap reduces as both networks converge, the pretrained UC-ffhq obtains a lower FID score after 2m images.
The PC-x strategy results are more intriguing.
The PC-concat-scratch model outperforms PC-add-scratch in terms of FID scores. 
This shows that pose information is fused better with the model using concatenation vs addition operator.
The pretrained PC-add-ffhq model gets lower FID scores initially but is outperformed by the PC-concat-scratch model eventually.
The PC-add-ffhq takes 40 hrs to train (has pretrained weights and fewer channels) while PC-concat takes 58 hrs.

We observe that both PC-x and UC-x models lead to similar FID scores and appear similar perceptually. While UC-x models generate arbitrary human poses, with PC-x models, we can control the pose of the output images. 
We also attempt initializing and training the PC-x models with the UC-x trained model — although FID scores drop quickly, the input-output poses are not aligned and require training the pose encoder further.

\noindent {\bf Probing the Latent Space.} The space of latent vectors $W$ embeds both style and pose information and controls the generated images.
We can obtain the latent vectors of an image by an {\it inversion} optimization process~\cite{stylegan2,togan}.
For a pair of images, we can {\it interpolate} between the two latent codes, to obtain new images with mixed style and pose information from $L_i$:
$L_i = L_{start} + i\cdot(L_{end} - L_{start})/{num\_latents}$.
We investigate two properties of the latent codes: (a) whether interpolated images exhibit {\it continuity} in style and pose, and (b) can pose and style codes be {\it disentangled}?

\begin{figure}[h]
\centering
\includegraphics[width=1.0\textwidth]{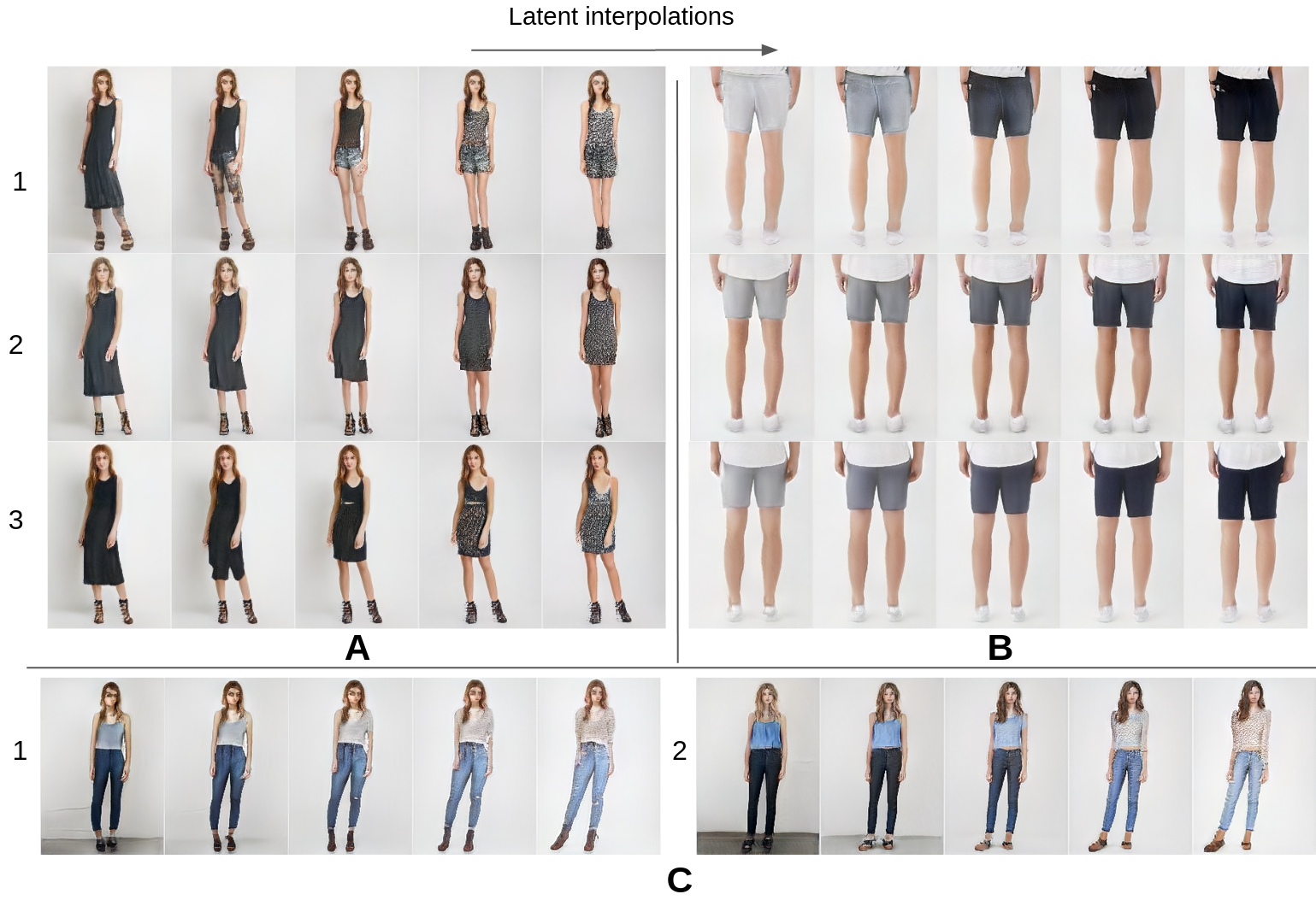}
\caption{Interpolation (A) Garment length (B) Color (C) Pose. 1. UC-ffhq 2. UC-scratch 3. PC-concat-scratch.}
\label{fig:interpolation}
\end{figure}

\begin{figure}[h]
\centering
\includegraphics[width=0.5\textwidth]{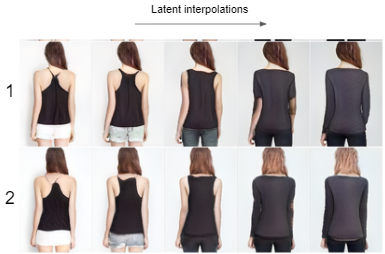}
\caption{Interpolation of garment region 1. UC-scratch 2. PC-add-scratch}
\label{fig:region_interp}
\end{figure}

To check {\it style continuity}, we select pairs of DeepFashion images that are similar in pose but different in garment color or shape, and interpolate between them.
Fig.~\ref{fig:interpolation}A shows interpolation results between images with different garment length.
Models trained from scratch (x-scratch) generate results with garment length varying smoothly, whereas transfer learned model, UC-ffhq generates discontinuous results. 
This may be due to pretrained features being less useful and small fine-tuning dataset~\cite{deepfashion}.
Fig. ~\ref{fig:region_interp} shows interpolation results between images with different garment region. The intermediate images are realistic images with novel dresses not present in any of the original images.
As Fig.~\ref{fig:interpolation}B shows, all models generate {\it color-continuous} interpolated images.

Fig.~\ref{fig:interpolation}C shows interpolation results from UC-x models on images which differ in pose, but similar in style.
Interpolation for small differences in pose are gradual and continuous, but between very different poses are not meaningful. 
Interpolation between poses may be helpful if we want to generate a variety of poses with the same garment and intermediate key-points are not available.

\begin{figure}[h]
\centering
\includegraphics[width=1.0\textwidth]{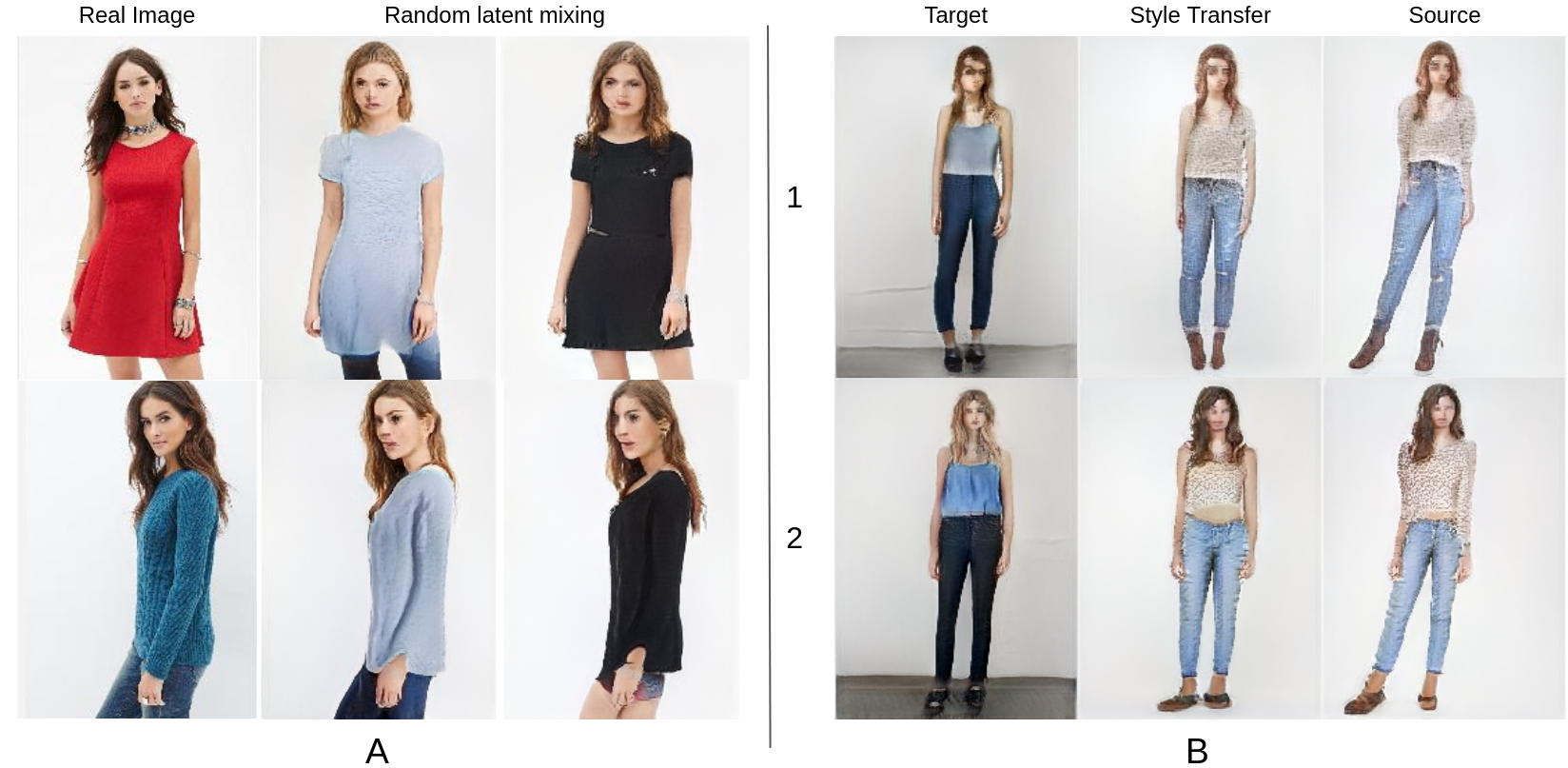}
\caption{Style latent mixing 1. UC-ffhq 2. UC-scratch.}
\label{fig:latent_mixing}
\end{figure}

The latent codes in the UC-x models (without pose encoder) do not disentangle pose and style during training.
Surprisingly, we observe self-emerging disentanglement: the {\it lower} (6 style blocks inputs) latent features determine the {\it pose} primarily while the {\it rest} (8 blocks) determine the {\it style}.
Fig.~\ref{fig:latent_mixing}A shows images generated by mixing latent codes of real images with randomly generated latent codes. Images are chosen randomly in this case.
Note that the pose in the generated images is same as the real image but the style is different. 
Interestingly, this observation enables a simple try-on transfer method with the UC model.
After inverting both the target image $P$ and the source image $S$, we can combine the lower latent codes from $P$ and higher ones from $S$, to generate the desired try-on image.
Fig.~\ref{fig:latent_mixing}B shows images generated with this method.


\section{Conclusions}
In this paper, we explored how transfer learning and pose conditioning affect the TryOnGAN model and investigated different properties of latent space interpolation.
In the ongoing work, we will explore how to generate higher quality images efficiently, new techniques for latent space interpolation, and enabling further disentanglement of pose and garment properties.

\bibliographystyle{ACM-Reference-Format}
\bibliography{togan}
\end{document}